\documentclass[3p, review, 11pt]{elsarticle}

\usepackage{flushend}
\usepackage{epsfig}
\usepackage{amsmath}
\usepackage{flushend}
\usepackage{multirow}
\usepackage{color}
\usepackage{enumerate}
\usepackage{easylist}
\usepackage{url}

\usepackage{amsthm}
\usepackage{amssymb }
\usepackage{amsmath}
\usepackage{bm}
\usepackage{float}
\usepackage{mathrsfs}
\usepackage{amsfonts}
\usepackage{graphicx}
\usepackage{algorithm, algpseudocode}
\usepackage{subfigure}
\def\SEANet{\text{BP-Net}}
\def\BPNet{\text{BP-Net}}
\def\NX{N_x}
\def\NS{N_s}
\def\l{^{(l)}}
\def\x{\bm{x}}
\def\p{\bm{p}}
\def\b{\bm{b}}

\def\h{\bm{h}}

\def\t{(t)}
\def\x{\bm{x}}

\algnewcommand\Input{\item[\hspace{6pt}\textbf{Input:}]}
\algnewcommand\Output{\item[\hspace{6pt}\textbf{Output:}]}
\algnewcommand\OutputVal{\textbf{output} }
\begin{document}

\begin{frontmatter}

\title{$\SEANet$: Cuff-less, Calibration-free, and Non-invasive Blood Pressure Estimation via a Generic Deep Convolutional Architecture}

\author{Soheil Zabihi$^\ddag$, Elahe Rahimian$^\dag$, Fatemeh Marefat$^\top$,  Amir Asif$^\ddag$, Pedram Mohseni$^\top$, and Arash Mohammadi$^\dag$}

\address{$\ddag$ Electrical and Computer Engineering, Concordia University, \\
1455 De Maisonneuve Blv. W., EV-009.187, Montreal, QC, Canada, H3G-1M8\\
$\dag$Concordia Institute for Information Systems Engineering,  Concordia University\\
$^\top$ Electrical, Computer, and Systems Engineering, Case Western Reserve University, Cleveland, OH, USA\\
Emails:' $\{$s\_zab, arashmoh, e\_ahimia$\}$@encs.concordia.ca}

\begin{abstract}
\textit{Objective}: The paper focuses on development of robust and accurate processing solutions for continuous and cuff-less blood pressure (BP) monitoring. In this regard, a robust  deep learning-based framework is proposed for computation of low latency, continuous, and calibration-free upper and lower bounds on the systolic and diastolic BP. \textit{Method}: Referred to as the $\SEANet$, the proposed framework is a novel convolutional architecture that provides longer effective memory while achieving superior performance due to incorporation of casual dialated convolutions and residual connections. To utilize the real potential of deep learning in extraction of intrinsic features (deep features) and enhance the long-term robustness, the $\SEANet$ uses raw Electrocardiograph (ECG) and Photoplethysmograph (PPG) signals without extraction of any form of hand-crafted features as it is common in existing solutions. \textit{Results}: By capitalizing on the fact that datasets used in recent literature are not unified and properly defined, a benchmark dataset is constructed from the MIMIC-I and MIMIC-III databases obtained from PhysioNet. The proposed $\SEANet$ is evaluated based on this benchmark dataset demonstrating promising performance and shows superior generalizable capacity. \textit{Conclusion}: The proposed $\SEANet$ architecture is more accurate than canonical recurrent networks and enhances the long-term robustness of the BP estimation task. \textit{Significance}: The proposed $\SEANet$ architecture addresses key drawbacks of existing BP estimation solutions, i.e., relying heavily on extraction of hand-crafted features, such as pulse arrival time (PAT), and; Lack of robustness. Finally, the constructed $\SEANet$ dataset provides a unified base for evaluation and comparison of deep learning-based BP estimation algorithms.
\end{abstract}


\end{frontmatter}
\section{Introduction} \label{sec:Introduction}
An alarming population ageing is widely expected in near future partially due to recent advancements of biomedical health technologies. According to a recent publication by the United Nation~\cite{UN:2017}, the number of seniors over the age of $60$ is expected to double by $2050$, even it is projected that population of seniors will be more than population of minors/youth at ages $10$-$24$ by $2050$. Consequent of this inevitable worldwide population aging trend is significant increase in age-related health issues, in particular cardiovascular conditions. According to World Health Organization report and in a global scale, cardiovascular diseases account for approximately $17$ million loss of lives annually, which accounts for one third of the total deaths around the world. Of these, complications of hypertension account for $9.4$ million loss of lives annually. These facts call for an urgent quest to develop advanced continuous monitoring, efficient diagnosis, and timely treatment of cardiovascular conditions. The paper focuses on the former category, i.e., continuous monitoring of Blood Pressure (BP), which is a widely measured physiological characteristic used to indicate cardiovascular health status of an individual.

Generally speaking, BP can be described as the pressure applied by blood to the arteries (wall of the blood vessels) ranging between two limits, i.e., a maximum value named Systolic Blood Pressure (SBP) to a minimum value referred to as the Diastolic Blood Pressure (DBP). A person is diagnosed by hypertension when her/his SBP goes above $140$mmHg and the DBP reaches above $90$ mmHg. The BP, however, varies significantly over time~\cite{6} as the result of different environmental factors such as stress, mental situations, and food consumption to name but a few. This necessitates continuous monitoring of BP to assist in early detection and diagnosis of hypertension conditions together with devising accurate treatments to reduce overall mortality rate of patients suffering from hypertension. While regular BP checkups are recommended by physicians for seniors, this typically can not be achieved due to complications of human activities and fast pase of modern life style, furthermore, historical data shows that on average $20$\% of seniors have higher measured BP at clinics in comparison to relaxed home environment. Consequently, continuous and in-home monitoring of BP~\cite{P2_2021}\nocite{P3_2020, Mukkamala:2018-2}-\cite{Miao:2019} via utilization of advanced Biological Signal Processing (BSP), Artificial Intelligence (AI) and Machine Learning (ML) techniques~\cite{P1_2020}\nocite{Qiu:2021, P5_2019, Zhang:2019, Khalid:2018}-\cite{Simjanoska:2018} becomes of paramount importance and is the focus of the paper. In Section~\ref{Sec:RelW}, we provide an overview of recent literature in this domain to better position the contributions of the paper (outlined below), and motivate the need for development of the proposed $\SEANet$ architecture.

\vspace{.025in}
\noindent
\textbf{Contributions}: The paper proposes a novel deep learning framework, referred to as the $\SEANet$, that addresses the identified drawbacks of existing solutions. More specifically, we propose a novel convolutional architecture for estimating BP. After denoising the photo-plethysmograph (PPG) and electrocardiogram (ECG) signals~\cite{P6_2018}, the pre-processed signals are provided as inputs to the designed convolutional architecture, i.e., excluding the need for feeding the models with hand-crafted features, therefore, utilizing the intrinsic deep features of the PPG and ECG signals. In brief, the main contributions of the paper can be summarized as~follows:
\begin{itemize}
\item Most of the existing data-driven methodologies proposed for cuff-less BP estimation depend on extraction of specific hand-crafted features such as  pulse arrival time (PAT)~\cite{Xiao:2018}\nocite{8-PWV, 6-PTT, 7-PTT, 8-PTT, 9-PTT, Yoon: 2018, Tang:2017}-\cite{6(12)}. Capitalizing on recent evidence that neural networks can extract the necessary features automatically without the need for complex feature engineering, we propose a convolutional architecture model that extracts necessary features automatically using raw ECG and PPG waveforms. The proposed model is able to estimate BP with high accuracy in an end-to-end manner.
\item The proposed $\SEANet$ provides longer effective memory while achieving superior performance in comparison to recurrent neural networks due to incorporation of casual dialated convolutions and residual connections.
\item In the studies presented so far, the type of ECG signal is not considered (i.e., the BP is estimated based on one type of ECG). Availability of  different types of ECG signals (which is not the same for every subject) makes it difficult to evaluate the generality of the obtained results. To address this problem, and also to show the generality of the proposed model, we use different types of ECG signals such as I, II, III, and IV.
\item By capitalizing on the significant importance of continuous BP monitoring and the fact that datasets used in recent literature are not unified and properly defined, a benchmark data set is constructed from the MIMIC-I and MIMIC-III databases  from PhysioNet to provide a unified base for evaluation and comparison of deep learning-based BP estimation algorithms. 
\end{itemize}
The rest of the paper is organized as follows: Section~\ref{Sec:RelW} outlines recent continuous BP estimation solutions and motivates the need for development of the proposed $\SEANet$ architecture. The proposed $\SEANet$ architecture is then presented in Section~\ref{sec:Framework}. Section~\ref{sec:Results} presents the experimental results for the evaluation of the proposed framework based on real constructed datasets. Finally, Section~\ref{sec:Conclusion} concludes the paper.

\section{Related Works}\label{Sec:RelW}

Generally speaking, when it comes to measuring BP, commonly, either a cuff-based approach is utilized, which provides upper arm BP measurements, or one resorts to cuff-less (possibly invasive) solutions. Upper arm BP monitoring can provide users with an indirect and non-continuous BP measurement technique by using an inflatable cuff and stethoscope. Such cuff-based methods suffer from several drawbacks including: (i) Being inconvenient and unhealthy especially in  public places; (ii) Require proper training prior to utilization; (iii)~Not being ideal for self-use and long-term monitoring of BP, and; (iv) Being incapable of providing continuous BP measurements~\cite{6}. Cuff-less BP monitoring~\cite{5}, on the other hand, eliminates the common uncomfortable factors associated with the former category and has the potential to continuously provide BP estimates without using any inflatable cuff.

Recently, there has been a great surge of interest towards the goal of performing continuous BP monitoring via \textit{Physiologically Inspired Models}~\cite{Xiao:2018}\nocite{8-PWV, 8-PTT}-\cite{9-PTT} in particular pulse transit time (PTT) and pulse arrival time (PAT). The PAT, sum of PTT and pre-ejection period, is considered as the main marker of BP for development of cuff-less BP estimation algorithms due to its simple measurement procedure. The PAT~\cite{6-PTT, 7-PTT} is defined as the time required for a heart beat to transfer to a body peripheral and has a tight relationship (correlation) to the BP. The existing correlation between PAT and BP, although well established, is highly non-linear depending on several uncertain factors varying across different individuals and over time~\cite{Yoon: 2018, Tang:2017}. Therefore, there has been different attempts~\cite{8-PTT}\nocite{9-PTT}-\cite{6(12)} to construct calibration models to account for such variations.   In calibration-based models, the focus is on extracting meaningful features to be fed to processing and learning models. Such models, however, are applicable only for use in short intervals such as exercise tests. Existing methods for cuff-less and continuous BP estimation can be classified into the following two main categories:

\vspace{.025in}
\noindent
\textit{\textbf{(i) Hand-crafted Regression-based Models}}: Models belonging to this category are developed by extracting hand-crafted features and exploiting various conventional BSP and ML algorithms such as decision trees (DT), support vector regression (SVR), shallow neural networks, and Bayesian linear regression (BLR) to name but a few. Typically, PPG and ECG signals are used jointly~\cite{6, Sharifi:2019} to extract PAT features to construct a regression model for estimating the BP. In Reference~\cite{3}, for instance, a linear regression model (i.e., the SVR) is coupled with a radial basis function (RBF) kernel, and a single hidden layer neural network with a linear output to estimate the BP. 

\textit{The key drawback of the aforementioned conventional methods is that such models rely heavily on extraction of hand-crafted features, such as PAT, and directly map the given input into the target value while ignoring the critical temporal dependencies in the BP dynamics.} This could be considered as the root of long-term inaccuracy of such models, and results in lack of robustness due to strong dependencies on the calibration parameters and the choice of hand-crafted features to describe the signal for subsequent regression~\cite{11}. Lack of robustness translates into accuracy decay over time requiring frequent calibrations, especially for multi-day continuous BP prediction tasks. Robustness is essential for use of developed models in clinical settings, therefore, lack of robustness of existing solutions has limited their practical utilization.

\vspace{.025in}
\noindent
\textit{\textbf{(ii) Deep Learning-based Models}}: While research works on hand-crafted and regression-based models are extensive, deep-learning based BP estimation~\cite{9,14} is still in its infancy. In deep-learning models, commonly, hand-crafted features (e.g., extracted PAT features) are fed to neural network models such as  long short term memory (LSTM) models, recurrent neural networks (RNN), convolutional neural networks (CNN), or bidirectional RNN (BRNN). For instance, Reference~\cite{9} proposed to formulate BP prediction as a sequence learning problem, and proposed a deep RNN model, which is targeted for multi-day continuous BP prediction. The RNN model works with a set of seven representative hand-crafted features extracted from ECG and PPG signals. Another recent example is Reference~\cite{10}, where the authors proposed a waveform-based Artificial neural network (ANN)-LSTM model. The model consists of a hierarchical structure where the lower hierarchy level uses ANNs to extract the required features from the ECG/PPG waveforms, while the upper hierarchy level uses stacked LSTM layers to learn the time domain variations of the features extracted in the lower hierarchy level.

One can identify two key drawbacks associated with the limited deep learning-based models developed recently for BP estimation: (a) In most of the studies presented so far,  before extracting the deep features, representative hand-crafted features of input signals are first selected/extracted, which are then used to train a deep neural network. In other words, such methods ignore the real potential of deep learning in utilizing the intrinsic features (deep features) of the input signals, and; (b) Lack of a benchmark dataset for evaluation and comparison of developed deep learning-based BP estimation algorithms. In other words, datasets used in recent literature are not unified and properly defined, which makes evaluations and comparison difficult. Commonly a subset of  MIMIC-I or  MIMIC-III databases from PhysioNet is used without providing details on the training, validation, and test sets rendering reproducibility and fair comparisons impossible. In some cases, results evaluated based on two different datasets are compared, which can not be considered as a fair base for evaluation purposes. Furthermore, commonly a limited number of subjects within specific age/gender group is used making it difficult to evaluate the generality of the obtained results across different characteristics.

To alleviate the aforementioned identified shortcomings of existing continuous BP estimation solutions, next we present the proposed $\SEANet$ architecture.
\section{The $\SEANet$ Framework}\label{sec:Framework}
\begin{figure}[t!]
\centering
\includegraphics[scale=.35]{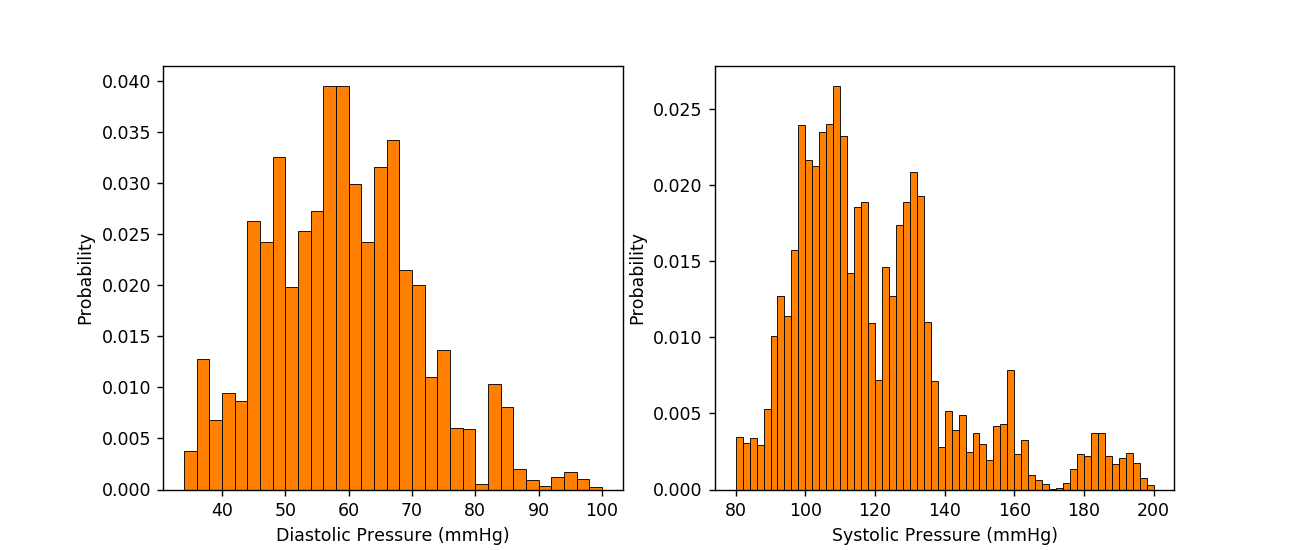}
\vspace{-0.1in}
\caption{\small Histogram of the preprocessed database. (a) Diastolic blood pressure. (b) Systolic blood pressure.\label{data_disterbution}}
\end{figure}
In this paper, estimation of the SBP and the DBP is performed automatically via extraction of deep-features from ECG and PPG signals without incorporation of any form of hand-crafted PAT features. To achieve this goal, we approach the BP estimation problem as a sequence modelling task. Before describing the architecture of the proposed $\SEANet$, in what follows, first we provide a brief overview of the constructed dataset and pre-processing pipelines utilized to develop the proposed $\SEANet$.

\subsection{$\SEANet$ Dataset}
As stated previously, the paper introduces a unified and properly defined benchmark dataset given the significant importance of continuous BP monitoring and the fact that recent research works train and test their algorithms on  datasets of their choice, impeding a fair judgment between their solutions. Capitalizing on this issue, we aim to provide a platform with reference dataset, where different algorithms can be evaluated and compared by utilizing the same training set to optimize new processing algorithms, and the same test dataset to be used to measure the associated performance. Despite the differences, all existing studies share a common validation procedure in which experiments are conducted on a proprietary database containing fewer number of subjects with reference to our work, therefore, it is difficult to evaluate the generality of the obtained results across gender and age. For this purpose, we increased the number of subject ($104$ individuals), and show that this increase in the number of subjects with respect to prior works can lead to challenging issues in terms of classification accuracy and generality across subjects.

The $\SEANet$ dataset is collected from the Multi-parameter Intelligent Monitoring for Intensive Care (MIMIC)~\cite{MIMIC_I} provided by PhysioNet server. MIMIC-I database contains PPG, multi-lead ECGs, and arterial blood pressure (ABP) signals at $125$ samples per second with $8$ bit precision. Data is derived from $90$ patients monitors in the medical, surgical, and cardiac intensive car units (ICU) of different hospitals~\cite{MIMIC_I}. The requirement set froward to construct the $\SEANet$ dataset is to have concurrent PPG, ECG, and ABP signals, therefore, out of the $90$ available subjects, we were able to collect data from $56$ patients. To further increase the number of subjects, MIMIC-III~\cite{MIMIC_I} database, which is an update to the common MIMIC-II, is also used as the second source for data preparation. The MIMIC-III contains data associated with a large number of different hospitals for distinct patients in ICU between $2001$ and $2012$. Similar to MIMIC-I, signals were sampled at the frequency rate of $125$ Hz with $8$ bit accuracy~\cite{MIMIC_I}. We have collected data from $48$ patients who had simultaneous PPG, ECG, and ABP from this dataset resulting in total of $104$ subjects in the $\SEANet$ dataset.  The BP distribution histograms are shown in Fig.~\ref{data_disterbution}. After preparing the input and output for the model, the data for each patient is divided into $70$\% for training, $10$\% for validation and $20$\% for test. The constructed $\SEANet$ dataset including train, validation and test subsets are available through its web-page: \url{http://i-sip.encs.concordia.ca/datasets.html}.

\subsection{Preprocessing}
\begin{figure}[t!]
\vspace{-0.1in}
\centering
\includegraphics[scale=.75]{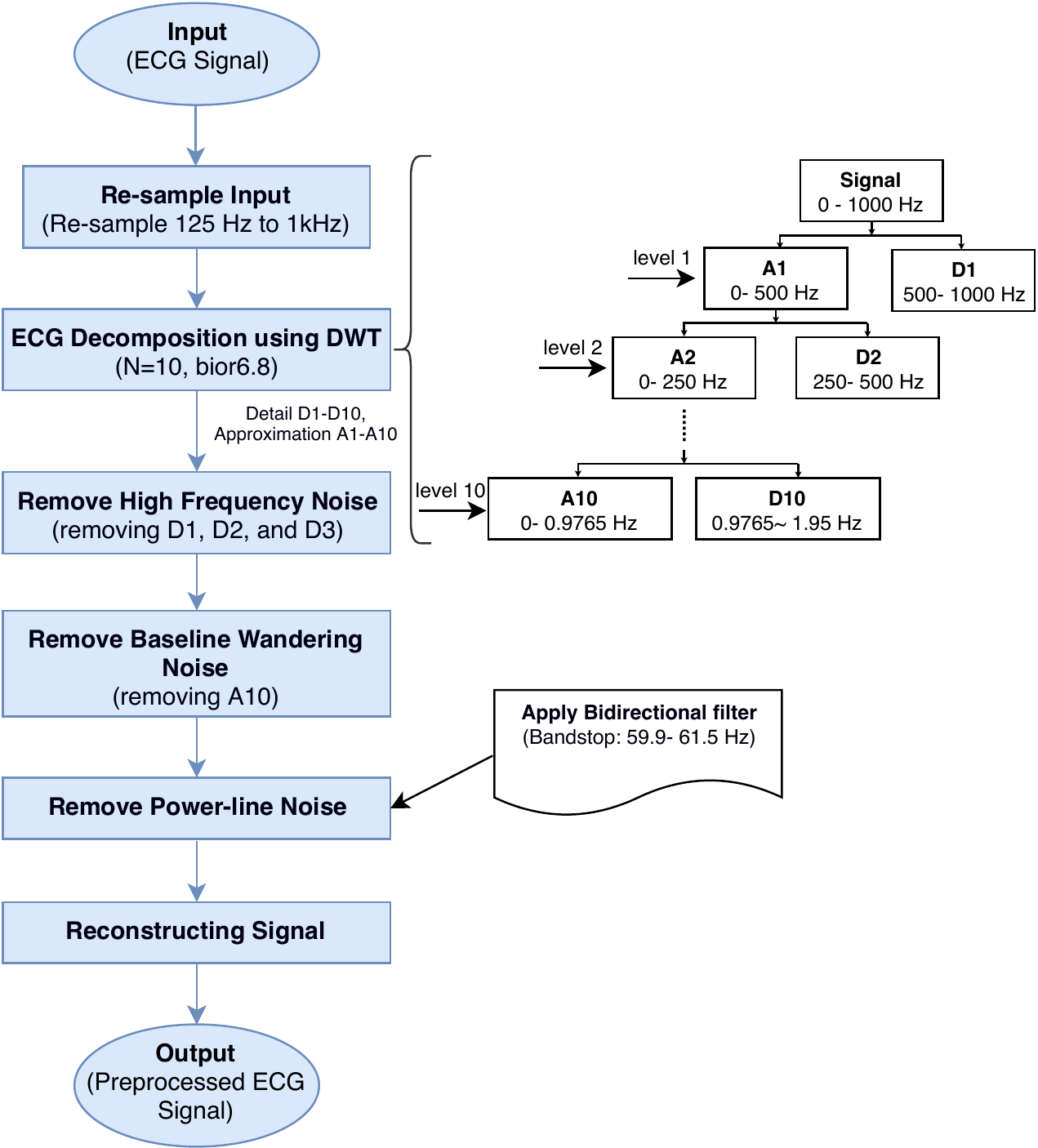}
\caption{\small Raw signal pre-processing pipeline.\label{Preprocessing_pipeline}}
\vspace{-.2in}
\end{figure}

Before describing details of the proposed $\SEANet$ for BP estimation, in this sub-section we present the essential pre-processing steps performed on ECG and PPG signals. Generally speaking, the ECG signals are contaminated by different noise and artifact sources~\cite{6}\nocite{13}-\cite{15} such as high frequency noise (e.g., electrosurgical noise, muscle contraction noise), power line interference noise ($50$ Hz or $60$ Hz), and low frequency noises (e.g., baseline wandering noise, motion artifacts affected by skin-electrodes). A great number of methods are introduced in the literature to remove noise/artifacts from ECG signals including Kalman filter-based methodologies~\cite{Kalman filter_removal_noise(3),  Kalman filter_removal_noise(4)}, filtering techniques (IIR or FIR filters), adaptive noise cancellation, empirical mode decomposition techniques, variational mode decomposition, and discrete wavelet decomposition (DWT)~\cite {survey_on_ECG denoising}. Although each of the such techniques has its pros and cons, in this work, the DWT technique~\cite{12} is applied to remove noise from signals since in contrary to other methods, it can be applied to non-stationary signals without introducing any artificial information to the original signal.

By implementing the pre-processing block shown in Fig.~\ref{Preprocessing_pipeline}, each ECG signal is denoised and prepared as an input for the proposed $\SEANet$ architecture. The first step for filtering ECG signals is up-sampling the frequency from $125$ Hz to $1,000$ Hz, which makes it stable to possible changes of the sampling frequency. Then, by applying \textit{Biorthogonal 6.8 (bior6.8)} mother wavelet with $10$ decomposition levels, the signal is decomposed into approximation and detail components denoted by $D1$-$D10$ and $A1$-$A10$ to refer to the detailed, and approximation components, respectively. The detailed coefficients $D1$, $D2$, and $D3$, contain high frequency noises such as muscle contraction~\cite{13}, which are discarded as the next step. Then Approximation coefficient $A10$ is eliminated because it contains low frequency noises, which are caused by respiration, and body movements (this noise can be seen as a sinusoidal component). After that, a \textit{bidirectional} filter is used to suppress the $60$Hz power-line interference noise using a bandstop filter ($59.5$ - $61.5$ Hz). Finally, the signal is reconstructed by Inverse DWT (IDWT). We applied an approach of similar nature to the above discussion to pre-process PPG signals and remove potential noisy contaminations.

\subsection{The $\SEANet$ Architecture}
We consider the problem of BP estimation from ECG and PPG signals collected from $\NS = 104$ number of subjects, where ECG, PPG, and ABP data associated with subject $l$, for ($1 \leq l \leq \NS$), each has total of $T\l$ number of samples. We define an ECG vector $\x\l\t = [X\l_1, \ldots, X\l_{t}]^T$ consisting of samples from ECG time-series collected from the $l^{\text{th}}$ subject from the starting time ($t = 1$) to time ($t \leq T\l$). Note that $\x\l(T\l)$ is a vector representing all ECG samples available for the $l^{\text{th}}$ subject. Similarly, we define a PPG vector $\p\l\t = [P\l_1, \ldots, P\l_{t}]^T$ representing PPG measurements collected from the $l^{\text{th}}$ individual upto and including time ($t \leq T\l$). Finally, vector $\b\l\t = [B\l_1, \ldots, B\l_{t}]^T$ represents the BP values from time instant $1$ to $t$.

The goal of the $\SEANet$ architecture is to learn a nonlinear function $\h(\x\l\t, \p\l\t)$ that takes as input ECG $\x\l\t $ and PPG $\p\l\t$ sequences and provides a predicted value $\hat{B}\t$ for the BP at time $t$. One limitation is that the predicted BP output, $\hat{B}\t$, at time step $t$ only depends on the input data obtained prior to time $t$, i.e., there is no leakage of information from future into the past. The target of  network is to minimize the cost function $\mathcal{L}\left(B\t , \h\left(\x\l\t, \p\l\t\right)\right)$ between all the results of the hypothesized functions $\h(\cdot)$ with ECG and PPG input sequences, and the actual output $B\t$.  In what follows, we describe different aspects of the proposed $\BPNet$ architecture.

\begin{figure}[t!]
\centering
\includegraphics[scale= 1.5]{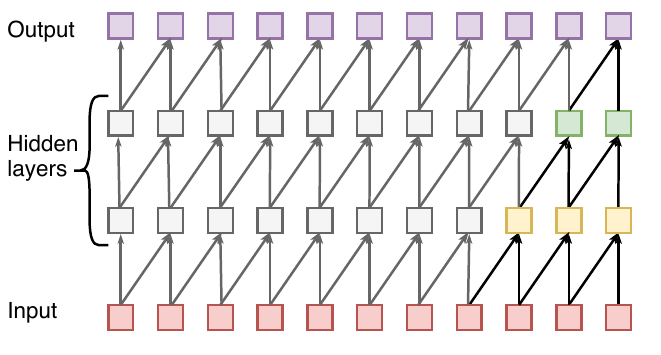}
\caption{An example of a Causal Convolution. \label{causal}}
\vspace{-0.2in}
\end{figure}

\vspace{.1in}
\noindent
\textbf{\textit{Casual Convolutions}}:
As stated above, a basic aspect of the proposed $\SEANet$ architecture is that we want to make sure that the output $\hat{B}\t$ estimated at time step $t$ depends only on previous input samples (i.e.,  ECG $\x\l\t $ and PPG $\p\l\t$ sequences) and not on any ``future'' values. In other words, while during the training phase, we have access to future values of the input signals (ECG and PPG sequences), a networked training using such an information can not be practically used to provide real-time predictions. To address this issue, one needs to implement causal filters within a deep learning architecture. The basic approach in this regard is to train the model without any restrictions on causality and during the implementation phase, mask those sections of the filter kernel that require future values as input. This masking approach can be achieved by setting the associated parts of the filter kernel to zero at each stochastic gradient decent (SGD) update. This approach is, however, costly in terms of required/wasted computational resources as, more-or-less, half of the multiplication and addition operations are wasted.  In the $\SEANet$ architecture we utilize an alternative approach, i.e., the \textit{Causal Convolutions}~\cite{wavenet} are incorporated within the $\SEANet$ architecture. Fig.~\ref{causal} illustrates one example of casual convolutions. In causal convolutions, by capitalizing on the translation-equivariance property of the convolution, the input signal is first shifted and padded by the kernel size and then the introduced shifting is removed.

For long sequences, architectures with causal convolutions are much faster to train than RNNs due to absence of  recurrent connections. However, one of the main drawbacks of the models with casual convolutions is limited receptive field. For example, in Fig.~\ref{causal}, the receptive field of the neurons at the output layer is only $4$, i.e., they see effects from upto four previous input samples (limited history size). For increasing the receptive field in casual convolutions, very deep networks or large filters should be applied, which are not generally feasible approaches. To address this issue, therefore, dilated convolutions~\cite{dilated} are used within the $\SEANet$ architecture as described below.

\begin{figure}[t!]
\vspace{-0.1in}
\centering
\includegraphics[scale=1.5]{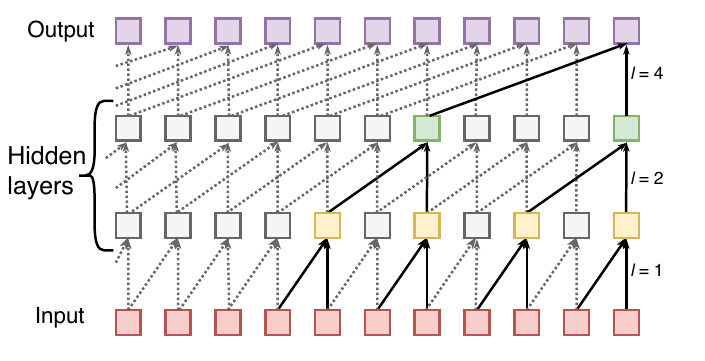}
\caption{\small An example of a Dilated Causal Convolution. \label{dilated}}
\vspace{-.2in}
\end{figure}

\vspace{.1in}
\noindent
\textbf{\textit{Dilated Convolutions}}:
Dilated convolutions are used as an effective way to enlarge the receptive field within the $\SEANet$ without losing resolution. Consider a 1-D time series $\x \in \mathbb{R}^{\NX}$ and a 1-D Kernel $\mathcal{K}: \{0, 1, \ldots, R-1\}\rightarrow \mathbb{R}$ with size $R$. Discrete dialated convolution operation $D(p)$ on the $p^{\text{th}}$ element of vector $\x$ with dilation rate $L$ is defined as follows
\begin{equation}
D(p) \triangleq (\x *_L \mathcal{K})(p) = \sum_{i = 0}^{R - 1} \mathcal{K}(i) \times \x(p - L\times i),
\end{equation}
where dialated convolution is denoted by $*_L$, and ($p-L\times i$) refers to elements of the input vector $\x$ prior to time $p^{\text{th}}$ element. Fig.~\ref{dilated} provides an illustration of dilated convolution. The dilation rate ($L$) is a design parameter, and dilated convolution becomes similar to the regular convolution with dilation rate of $L=1$. Generally speaking, in dilated convolution, the receptive field of a kernel $\mathcal{K}$ with size $R$ is expanded to  $R + (R-1)(L-1)$ with dilated stride of $L$.

\begin{figure}[t!]
\centering
\mbox{\hspace{-0.4in}\subfigure[]{\includegraphics[scale=1.5]{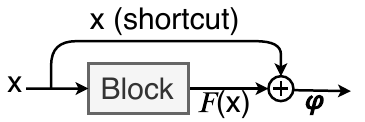}}\hspace{-0.03in}
\subfigure[]{\includegraphics[scale=1]{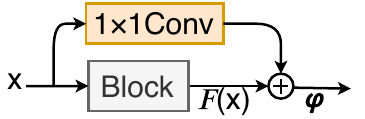}}}
\vspace{-.15in}
\caption{\small Residual learning (a) Identity block. (b) Convolutional block.
\label{identity}}
\end{figure}

\vspace{.1in}
\noindent
\textbf{\textit{Residual Connections}}
In theory, by stacking more layers, we expect that the network achieves lower training error and learn better, however,  in the reality, it is very challenging to train very deep neural networks because of vanishing and exploding gradient problems. In other words, by adding more layers in a deep neural network, accuracy gets saturated, and then degrades promptly. This problem is called ``degradation'', which was introduced by~\cite{resnet}. In ~\cite{resnet}, the authors, addressed the degradation problem by adding shortcut connections  to the network referred to as ``residual connections''. In this case, by skipping over some layers, information passes into the network. Fig.~\ref{identity}(a) shows a sample of ``Identity Block'' which is a standard block used in ResNets. This block is used when the dimensions of the input and the output are the same. In the case that the input and output dimensions do not match, another type of ResNets blocks called ``Resnet Convolutional Block'' are employed to match dimensions (done by 1$\times$1 convolutions)  as shown in Fig.~\ref{identity}(b). It is noteworthy to mention that in Fig.~\ref{identity}, $\varphi$ represents the activation function, which applied on the element-wise addition of residual mapping function ($F$) and the input ($x$).

\begin{figure}[t!]
\centering
\includegraphics[scale=1]{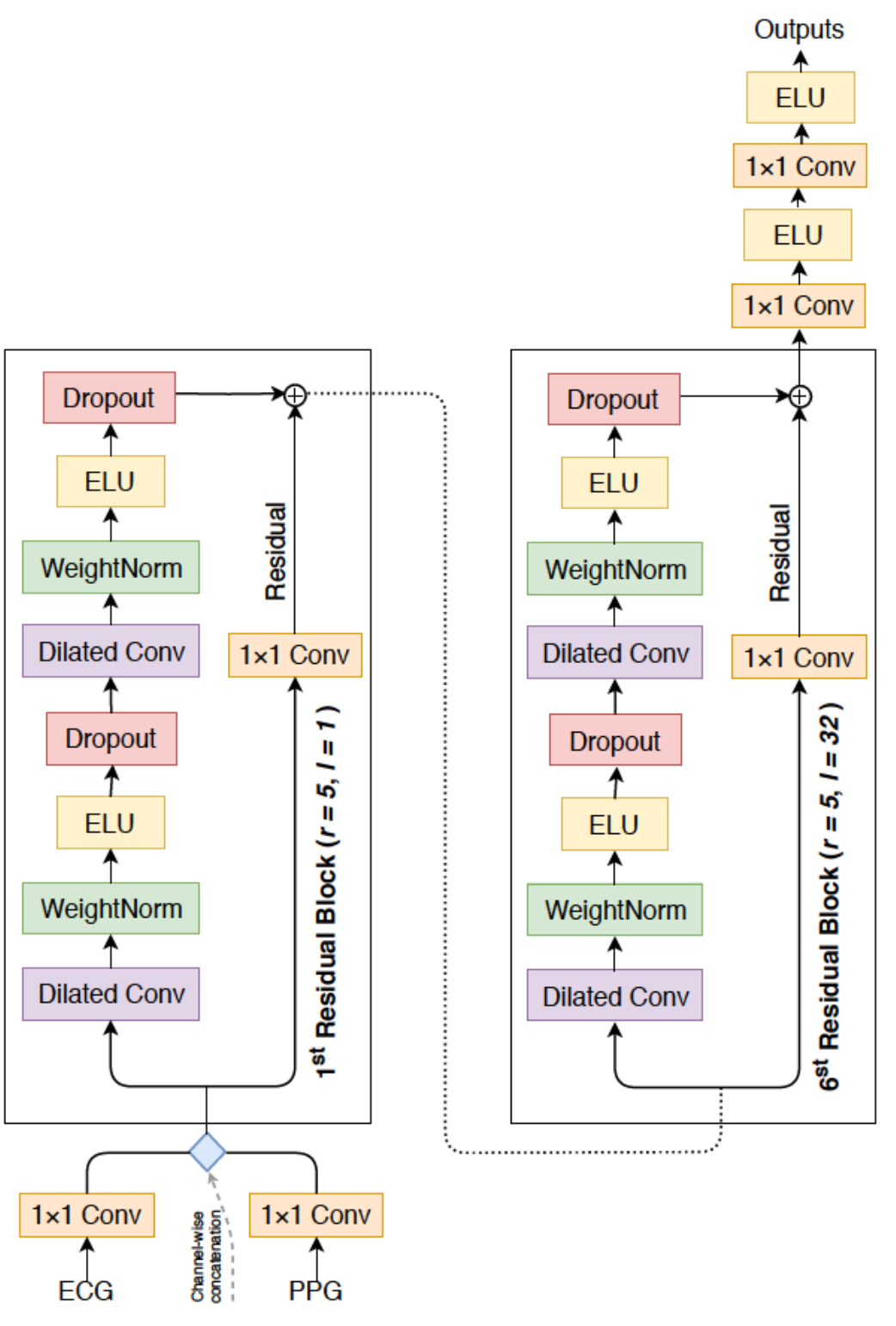}
\caption{\small The architecture of proposed $\SEANet$. \label{residual}}
\vspace{-.2in}
\end{figure}
\subsection{$\SEANet$ Structure and Hyperparameters Settings}
In the first step, the following nonlinear $\mu$-law transformation is applied to normalize the input data
\begin{equation}\label{mu_law}
F(X) = \text{sign}(X)\frac{\ln{(1+ \mu |X|)}}{\ln{(1+ \mu )}}.
\end{equation}
It is noteworthy to note that such $\mu$-law transformation is used as a transformation to quantize data in previous works such as~\cite{wavenet}. Inspired by Reference~\cite{TCN}, the residual block as a basic block is used for the network structure (Fig.~\ref{residual}). This block consists of two dilated casual convolutions and two nonlinear activation function (ELU)~\cite{Elu}. Moreover, dropout~\cite{dropout} for regularization, and weight normalization~\cite{weight}  are used in this block. We use Adam optimizer as the optimization algorithm with learning rate set to $0.001$. The learning rate changes in a cycle with the length of $100$ epochs. After $20$ epochs, we divide learning rate by $2$, but after $100$ epochs instead of dividing by $2$, we multiply it by $14.4$. Therefore, the learning rate at the beginning of each cycle will be $90$\% of the learning rate at the beginning of the previous cycle. This novel approach of creating learning rate causes to avoid being stuck in local minimums while speeding the training of the architecture. These models are trained with a mini-batch size of $64$. The exact structure of the proposed $\SEANet$ network is as follows:
\begin{itemize}
\item PPG and ECG signal as inputs are separately fed to a 1$\times$1 convolutions layer with $32$ kernels, and then the concatenated channel-wise results are fed to the first residual block.
\item Six residual blocks are stacked in the proposed architecture with the following characteristics:
\begin{itemize}
\item All of dilated causal convolutions have kernel size of $5$.
\item The dilation factor ($L$) is doubled for every layer, i.e., $1$, $2$, $4$, $8$, $16$, and $32$.
\item For Residual block $1$ to $6$, the number of kernels are $32$, $32$, $64$, $64$, $128$, and $256$,  respectively.
\end{itemize}
\item The output of the sixth block is fed to a 1$\times$1 convolution layer with $256$ kernels following by ELU activation function, ($1\times1$) convolution layer with $2$ kernels, and again an ELU activation function.
\end{itemize}
This completes presentation of the proposed $\SEANet$ architecture. Next, we present experimental results to evaluate performance of the proposed architecture.

\section{Experiments and Results}\label{sec:Results}
\begin{table}[t!]
\vspace{-0.1in}
\centering
\caption{\small The RMSE/MAE between actual BP(SBP, DBP) and estimated BP in the proposed model.\label{RMSE_t}}
\includegraphics[scale=1]{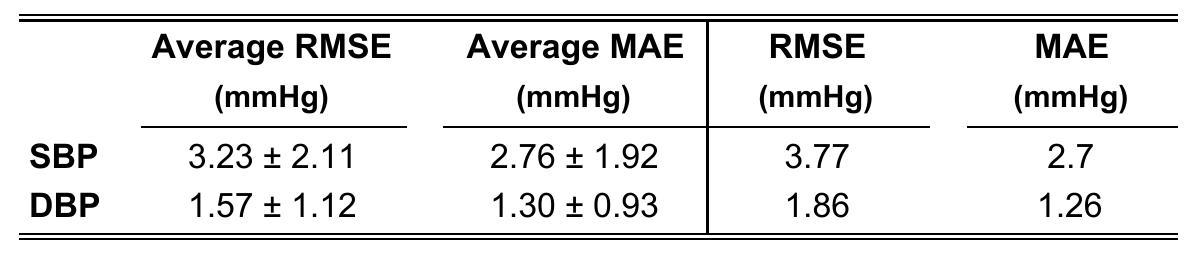}
\end{table}
\begin{figure}[t!]
\vspace{-0.1in}
\centering
\includegraphics[scale=.35]{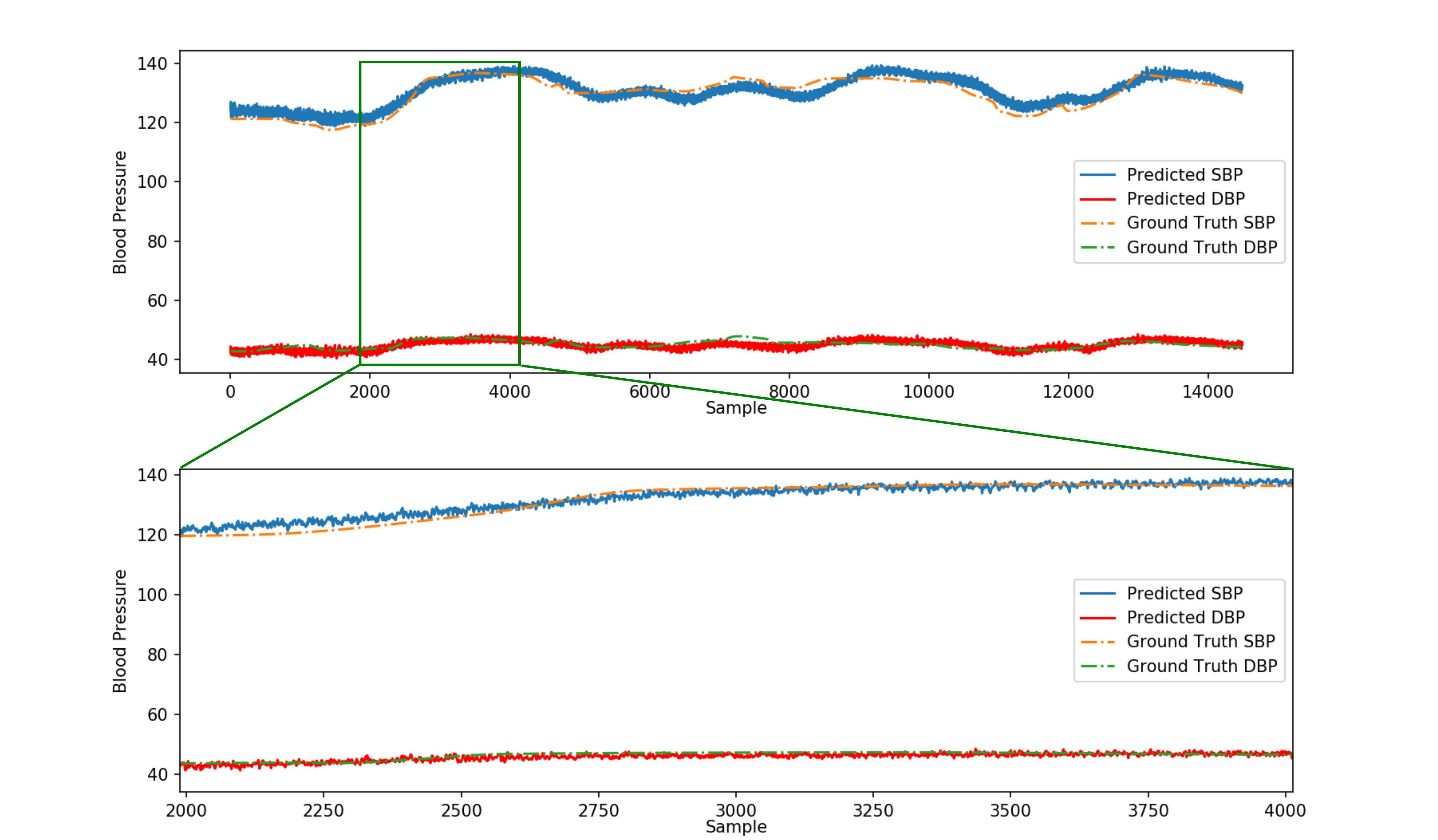}
\caption{\small Continuous BP measurement. \label{track_BP}}
\end{figure}
In this section, we present different experimental results based on real-data sets to evaluate performance of the proposed continuous BP estimation architecture. Details of the constructed dataset is described in Section~\ref{sec:Framework}. Fig.~\ref{track_BP} shows the continuous SBP and DBP tracking of an individual using the proposed $\SEANet$ architecture. From this figure, it is observed that the proposed model is capable of reliably  tracking continuous SBP and DBP.
To measure the accuracy of the proposed architecture, commonly used evaluation metrics, i.e., the root mean square error (RMSE) and the mean absolute error (MAE), are used given by
\begin{equation}
\text{RMSE} = \sqrt{\dfrac{1}{n}\sum\limits_{j=1}^{n}|B_{j} -\hat{B}_{j}|^2},
~\text{ and }~ \text{MAE} = \dfrac{1}{n}\sum_{j=1}^{n}|B_{j} - \hat{B}_{j}|,\label{RMSE}
\end{equation}
where $B_j$ is the actual observed BP, whereas $\hat{B}_j$ determines its corresponding predicted value. Table~\ref{RMSE_t}, summarizes the average RMSE and MAE results obtained from averaging individual RMSE and MAE values corresponding to each subject ($104$ subjects). Table~\ref{RMSE_t} also illustrates the combined MAE and RMSE results obtained by stacking together the target signals of all the subjected. The overall results presented in Table~\ref{RMSE_t} shows exceptional performance of the proposed  BP estimation framework especially for estimating the DBP.

\textit{It is worth mentioning that in contrary to existing BP estimation solutions, the proposed framework, provides continuous SBP and DBP outputs for all samples. Intuitively speaking, output of the proposed framework can be considered as upper and lower bounds on the BP signal computed at all time samples.}  As a side note to our discussion, we would like to point out that due to absence of a standard dataset, fair comparison with existing works is not straightforward.

\subsubsection{The BP Error Histogram Analysis}
\begin{figure}[t!]
\vspace{-0.1in}
\centering
\includegraphics[scale=.3]{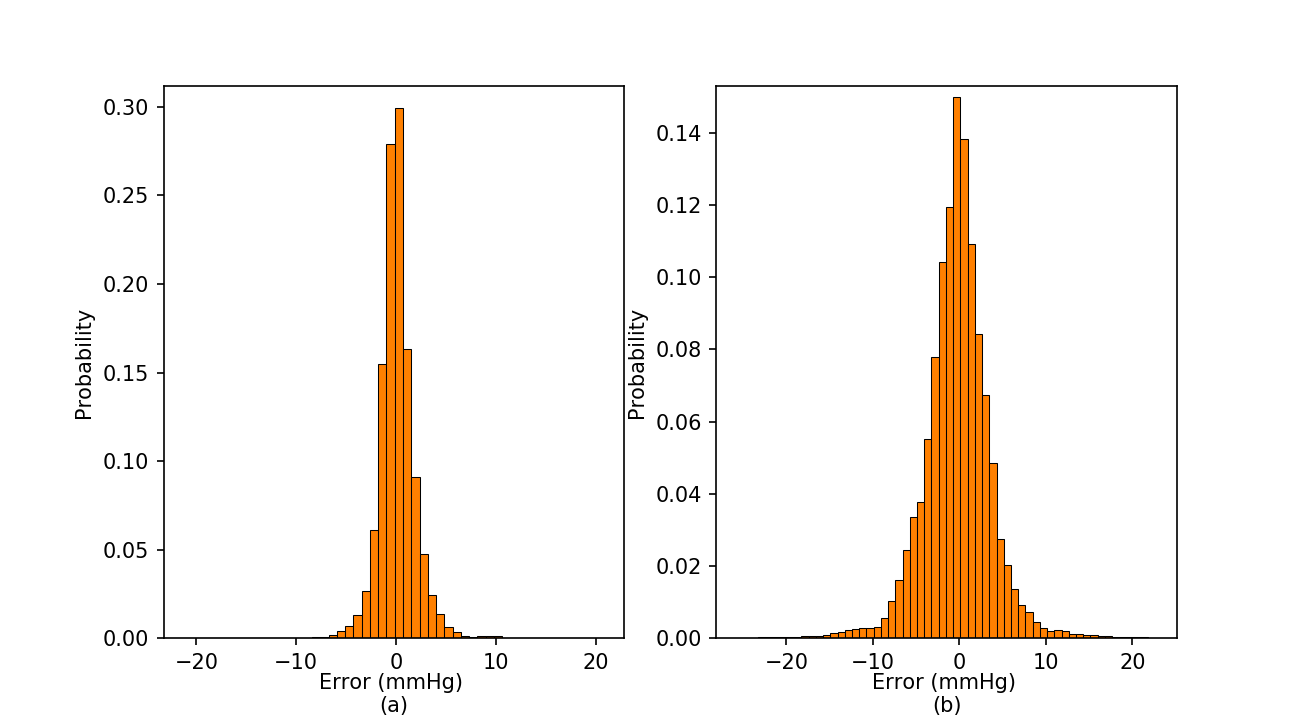}
\caption{\small (a) DBP error histogram of the proposed architecture. (b) SBP error histogram of the proposed architecture.
\label{error_histogram}}
\end{figure}
To determine reliability of the proposed $\SEANet$ architecture, histogram of the BP estimation error for DBP and SBP is shown in Fig.~\ref{error_histogram}(a)-(b). It is evident that the distribution of estimation error histogram can be modelled as a Gaussian function with approximately zero mean. As expected, standard deviation of the SBP error histogram is greater than that of the DBP, since the variance of the SBP target histogram is much larger than the DBP target histogram in database Fig.~\ref{data_disterbution}.

\subsection{Evaluation using AAMI Standards and BHS Standards}
\begin{table}[t!]
\vspace{-0.1in}
\centering
\caption{\small Comparison of the proposed model with the AAMI~standard.\label{AAMI_t}}
\includegraphics[scale=1]{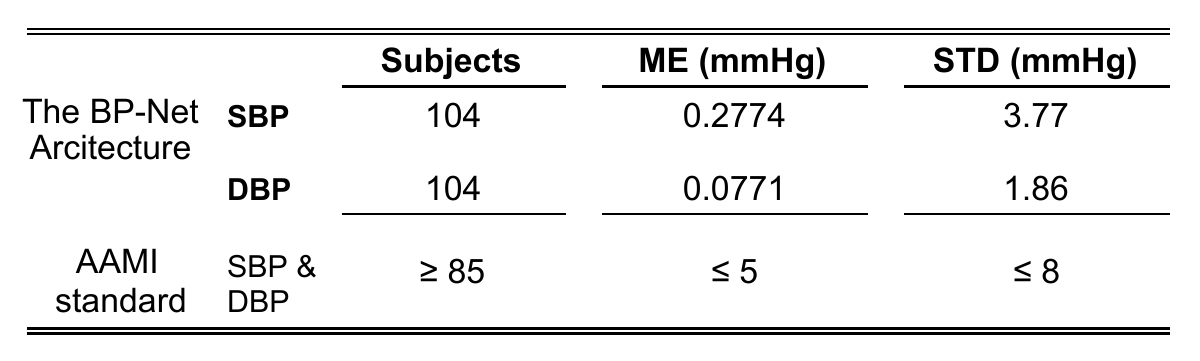}
\vspace{-.1in}
\end{table}
\begin{table}[t!]
\vspace{-0.1in}
\centering
\caption{\small Comparison of the proposed model with BHS standard.\label{BHS_t}}
\includegraphics[scale=1]{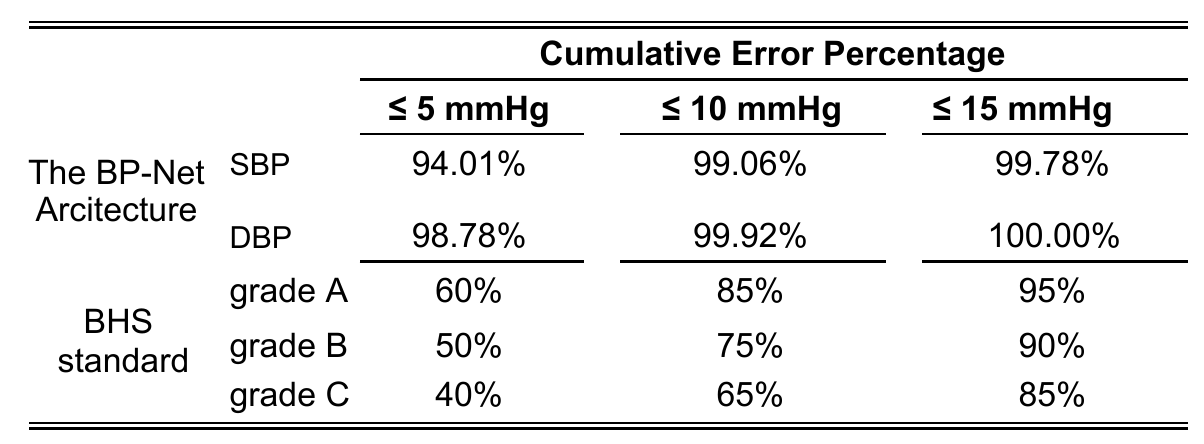}
\vspace{-.2in}
\end{table}

To evaluate robustness and accuracy of the proposed architecture, the model is evaluated on two common standards namely the Association for the Advancement of Medical Instrumentation (AAMI)~\cite{AAMI} and the British Hypertension Society Standard (BHS)~\cite{BHS}. Moreover, in the following subsections, we investigate  performance of the proposed $\SEANet$ architecture by using Bland-Altman test, and Pearson's correlation coefficient analysis.

According to the AAMI standard, the BP measurement devices should have mean error (ME) and standard deviation of error (SDE) values lower than $5$ and $8$ mmHg, respectively~\cite{AAMI}. Table~\ref{AAMI_t} shows the ME and SDE computed based on the overall results obtained from the proposed methodology. It can be observed that the proposed framework have ME and SDE values much lower than the acceptable ME and SDE requirements of the AAMI standard. The other requirement of the AAMI standard is that the device should be evaluated on a population more than $85$ subjects~\cite{6}, which is also satisfied in this work ($104$ subjects).

The BHS standard~\cite{BHS} grades BP measurement devices based on their cumulative percentage of errors lower than three different thresholds of $5$, $10$, and $15$ mmHg. Table~\ref{BHS_t} illustrates that not only the obtained results meet the requirements of the BHS standard, but also the grade of estimation for both DBP and SBP is equal to $A$ grade. Moreover, as shown in Table~\ref{BHS_t}, more than $94$\% of the estimated targets have cumulative error less than $5$ mmHg showing the exceptional performance of the proposed model.

\subsection{Statistical Analysis}
\begin{figure}[t!]
\vspace{-0.1in}
\centering
\includegraphics[scale=.3]{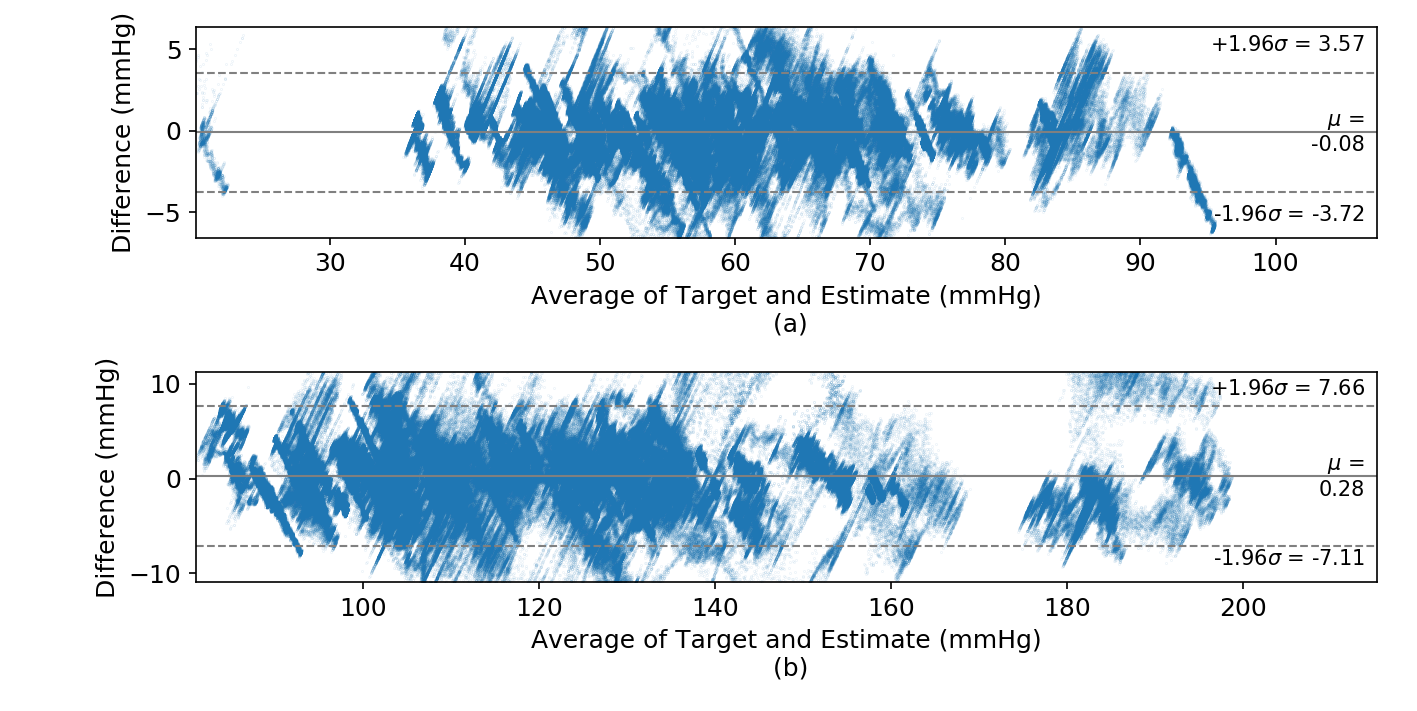}
\caption{\small Bland-Altman plot of (a) The DBP, and; (b) The SBP. The limits of agreement (LOA) for DBP and SBP are [$-3.72, 3.57$] and [$-7.11, 7.66$], respectively.
\label{bland_altman}}
\end{figure}
\begin{figure}[t!]
\vspace{-0.1in}
\centering
\includegraphics[scale=.35]{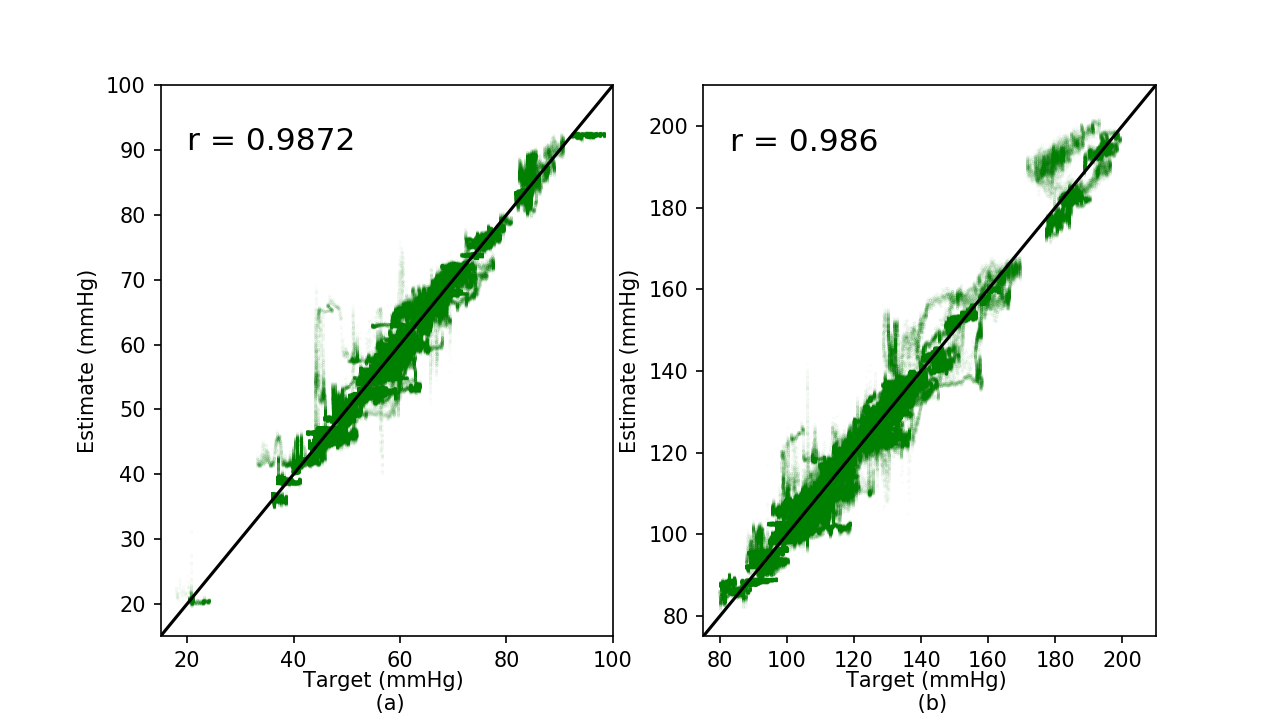}
\caption{\small The regression plot for (a) The DBP, and; (b) The SBP. The Pearson's correlation coefficients are $r = 0.9872$ and $r = 0.986$ for DBP and SBP, respectively.
\label{corrolation}}
\vspace{-.1in}
\end{figure}

Fig.~\ref{bland_altman} shows the Bland-Altman plot of the SBP and the DBP estimation. For the proposed $\SEANet$ architecture, the limits of agreement [$\mu$ - 1.96$\sigma $, $\mu$ + 1.96$\sigma $] for SBP and DBP have been found to be [$-7.11, 7.66$] and [$-3.72, 3.57$] respectively. This means that $95$\% of the estimated SBPs have error less than 7.73 mmHg and $95$\% of the measured DBPs have error less than $4.05$ mmHg, which indicates that the model provides acceptable estimates.

Finally, Figs.~\ref{corrolation}(a)-(b) illustrate the regression plot for SBP and DBP estimation. The Pearson correlation coefficients of DBPs and SBPs are $r = 0.9872$ and $r = 0.986$, respectively. Both of the coefficients are very close to $1.0$ indicating high linearity between the target and estimated BP.

\section{Conclusion}\label{sec:Conclusion}
Inevitable increase in the population of seniors, makes continuous BP monitoring essential as it provides invaluable information about individuals' cardiovascular conditions. The paper first identifies two key drawbacks associated with the existing continuous BP estimation models, i.e., (i) Relying heavily on extraction of hand-crafted features,  i.e., ignoring the real potential of deep learning in utilization of the intrinsic features (deep features) and instead using representative hand-crafted features  prior to extraction of deep features, and; (ii) Lack of a benchmark dataset for evaluation and comparison of developed deep learning-based BP estimation algorithms. To alleviate these issue, the paper proposes an efficient algorithm, referred to as the $\SEANet$, based on the deep learning techniques for the continuous, cuff-less, and calibration-free of systolic and diastolic BP. In the proposed $\SEANet$ architecture, raw ECG and PPG signals as utilized without extraction of PAT features, to explore the real potential of deep learning in utilization of intrinsic features (deep features). The proposed $\SEANet$ architecture is more accurate than canonical recurrent networks and enhances the long-term robustness of the BP estimation task.  Furthermore, by capitalizing on the significant importance of continuous BP monitoring and the fact that datasets used in recent literature are not unified and properly defined, a benchmark data set is constructed  from the MIMIC-I and MIMIC-III databases to provide a unified base for evaluation and comparison of deep learning-based BP estimation algorithms.  The proposed $\SEANet$ architecture is evaluated based on this benchmark data-set demonstrating promising results and shows generalizable capability to multiple subjects.


\end{document}